\documentclass{bmvc2k}
\usepackage{graphicx,import}
\usepackage{multirow}
\usepackage{algorithm}
\usepackage{algpseudocode}
\usepackage{amssymb}
\usepackage{makecell}
\usepackage{acro}
\usepackage{flafter} % place floats after reference
\usepackage{listings}

\usepackage[shortcuts]{extdash}

\DeclareAcronym{nn}{
    short = NN,
    long = Neural Networks,
}

\DeclareAcronym{qat}{
    short = QAT,
    long = Quantization\-/Aware Training,
}

\DeclareAcronym{bn}{
    short = BN,
    long = Batch Normalization,
}

\DeclareAcronym{ptq}{
    short = PTQ,
    long = Post-Training Quantization,
}

\DeclareAcronym{sota}{
    short = SOTA,
    long = state-of-the-art,
}

\title{RepQ: Generalizing Quantization-Aware Training for Re-Parametrized Architectures}

\addauthor{Anastasiia Prutianova}{auprutyanova@gmail.com}{1}
\addauthor{Alexey Zaytsev}{Likzet@gmail.com}{2}
\addauthor{Chung-Kuei Lee}{lee.chung.kuei@hisilicon.com}{1}
\addauthor{Fengyu Sun}{sunfengyu@hisilicon.com}{1}
\addauthor{Ivan Koryakovskiy}{i.koryakovskiy@gmail.com}{1}
\addinstitution{
 Huawei Technologies Co. Ltd.
 \\
 Shanghai, China
}
\addinstitution{
 Beijing Institute of Mathematical Sciences and Applications\\Beijing, China 
}
\runninghead{Prutianova, Zaytsev, Lee, Sun, Koryakovskiy}{RepQ}

\begin{document}

\maketitle

% Ivan's edits
\newcommand{\ik}[1]{\textcolor{magenta}{\parbox{\linewidth}{#1}}}

% add path for pdf_latex plots
\graphicspath{{images/}}

%-------------------------------------------------------------------------
\begin{abstract}

Existing neural networks are memory-consuming and computationally intensive, making deploying them challenging in resource-constrained environments. However, there are various methods to improve their efficiency.
Two such methods are quantization, a well-known approach for network compression, and re-parametrization, an emerging technique designed to improve model performance.
Although both techniques have been studied individually, there has been limited research on their simultaneous application. To address this gap, we propose a novel approach called RepQ, which applies quantization to re-parametrized networks.
Our method is based on the insight that the test stage weights of an arbitrary re-parametrized layer can be presented as a differentiable function of trainable parameters. We enable quantization-aware training by applying quantization on top of this function. RepQ generalizes well to various re-parametrized models and outperforms the baseline method LSQ quantization scheme in all experiments.

\end{abstract}
\section{Introduction}
\label{sec:intro}

The number of parameters in \ac{nn} has been growing over the years.
This substantial computational complexity precludes the deployment of real-life \ac{nn}-based applications to resource-constraint devices, e.g., mobile phones. 
Many research works aim at designing computationally efficient \ac{nn}.
A non-exhaustive list of ideas in this area includes knowledge distillation~\cite{hinton_distilling_2015}, model pruning~\cite{han_learning_2015}, matrix factorization~\cite{denil_predicting_2013, jaderberg_speeding_2014}, neural architecture search~\cite{zoph_neural_2017,liu_darts_2019}, quantization~\cite{li_ternary_2016, choi_pact_2018, esser_learned_2020}, and 
re-parametrization ~\cite{zagoruyko_diracnets_2017,ding_repvgg_2021}. 
Here, we focus on re-parametrization and quantization as our main research fields.

Re-parametrization is an emerging technique that recently allowed the training of a plain non-residual VGG~\cite{ding_repvgg_2021} model to achieve the remarkable accuracy of $80\%$ on ImageNet~\cite{russakovsky_imagenet_2014} while being faster than ResNet-101 \cite{he_deep_nodate}.
Moreover, re-parametrization has set a new \ac{sota} in channel pruning~\cite{ding_resrep_2021} and became a part of YOLO-7~\cite{wang_yolov7_2022}. 
The idea behind re-parametrization is that a neural architecture can be represented in different mathematical forms. 
Similar to how 2-D vectors on a plane can be defined by their Cartesian or polar coordinates, the same architecture can be expressed using various algebraic representations. 
An alternative representation helps gradient descent to reach a better local minimum, resulting in improved performance.
In practical applications, re-parametrization involves replacing linear layers, such as fully-connected and convolutional layers, with a combination of linear layers. 
For example, several papers~\cite{ding_acnet_2019, ding_repvgg_2021, guo_expandnets_2020, hu_online_2022, zagoruyko_diracnets_2017} have proposed replacing each convolution with a block consisting of multiple convolutional layers with various kernel sizes, channel numbers, residual connections, and batch normalization layers. 
This re-parametrization block is used during training but is equivalently converted back into a single convolution during inference.
To sum up, boosting model quality without additional computational burden at inference makes re-parametrization an important method for increasing \ac{nn} efficiency.
Efficiency is a feature of another major compression technique, namely quantization. Basic quantization algorithms can achieve more than a 75\% reduction in model size while maintaining a performance comparable to uncompressed models. 
This makes quantization a key technique for real-life model deployment.

We aim to bring the advances in re-parametrization research to practical applications by providing a suitable approach to quantizing re-parametrized \ac{nn}.
Currently, only two research works consider quantizing re-parametrized \ac{nn} \cite{ding_reparameterizing_2023,chu_make_2022}, and they both target a particular architecture, RepVGG.
Although novel, the obtained results show a considerable quality drop due to quantization which eliminates the improvements introduced by the re-parametrization.

To solve the problem of insufficient quality and generalization, we introduce a \ac{qat} strategy for re-parametrized \ac{nn}.
The challenge is that the training stage weights differ from the testing stage weights, prohibiting standard \ac{qat} application ~\cite{esser_learned_2020}.
In particular, applying a regular quantization independently to each convolution of a re-parametrized block prevents its merging during inference without increasing the quantizer bit width. For instance, two sequential convolutions of bit-width two will merge into a single convolution of bit-width four. 
Instead, we propose to compute the inference stage parameters of a convolution as a differentiable function of the trainable parameters of the re-parametrized block and then apply a pseudo-quantization function on top of it. 
This way, our RepQ approach enables end-to-end quantization-aware re-parametrized training.

To sum up, our main contributions are the following.
\begin{itemize}
\item We are the first to propose a method that allows \ac{qat} of models with arbitrary re-parametrization schemes. We provide extensive experiments showing that our RepQ method of joint quantization with re-parametrization leads to consistent quality enhancement on \ac{sota} architectures. In particular, for the first time, we achieved lossless 8-bit quantization of re-parametrized models.
\item \ac{bn} contained in the re-parametrized blocks creates a challenge for differentiable merging and hence for \ac{qat}. We show how to compute differentiable weight functions via BN-folding. Lastly, we enhance BN-folding
by introducing BN statistics estimation, which reduces the computational
complexity in theory and practice.

% \item We are the first to propose a method that allows \ac{qat} of models with diverse re-parametrization schemes. We show that joint re-parametrization and quantization-aware training is better than the sequential application of these methods. In particular, for the first time, we achieved lossless 8-bit quantization of re-parametrized models.
% \item \ac{bn} contained in the re-parametrized blocks creates a challenge for differentiable merging and hence for \ac{qat}. We propose several solutions for handling \ac{bn} in RepQ. 
% \item We provide extensive experiments on \ac{sota} re-parametrized architectures and show that our proposed method generalizes well for different re-parametrizated blocks.
\end{itemize}
\section{Related works}

% \subsection{Re-parametrization}

\textbf{Re-parametrization.} Early re-parametrizations appeared as a result of batch normalization~\cite{ioffe_batch_2015} and residual connection~\cite{he_deep_nodate} research. Inspired by batch normalization, the authors of~\cite{salimans_weight_2016} proposed 
a weights normalization technique. This technique can be viewed as a re-parametrization that decouples the direction and length of the weights vector by introducing an additional parameter responsible for the weight norm. The authors of DiracNets~\cite{zagoruyko_diracnets_2017} looked for a way to train deep networks without explicit residual connections. They proposed re-parametrizing a convolution with a convolution combined with a skip connection. It allowed very deep single-branch architectures to reach a decent quality. ACNets~\cite{ding_acnet_2019} and RepVGG~\cite{ding_repvgg_2021} brought re-parametrization to a new level by introducing multiple convolutions and batch normalization to re-parametrization blocks, revisiting well-known architectures like VGG~\cite{simonyan_very_2014} and ResNet~\cite{he_deep_nodate} and significantly enhancing their quality.
More advanced~\cite{ding_diverse_2021, hu_online_2022} re-parametrization strategies tend to use a larger number of convolutions and more diverse branches in their blocks.
Besides its extensive application to classification tasks, re-parametrization recently helped to reduce computational burden in other computer vision tasks like object detection~\cite{li_yolov6_2022, wang_yolov7_2022} and super-resolution~\cite{zhang_edge-oriented_2021,wang_repsr_2022}.
Re-parametrization receives theoretical justification in \cite{arora_optimization_2018}; the authors prove that under certain theoretical assumptions in a simple convex problem, re-parametrization leads to faster convergence. The authors of~\cite{ding_reparameterizing_2023} show that in some cases, re-parametrization can be equivalent to regular network training with a certain gradient scaling strategy.

% \subsection{Quantization}
\textbf{Quantization.} Regularly, NN parameters are stored as floating
point numbers. Yet, a 32-bit parameter representation is redundant to maintaining the network's quality. The research field aiming to find the optimal low-bit integer parameters' representation is known as neural network quantization.
Theoretically,  quantization involves rounding, leading to zero gradients in the network almost everywhere. 
QAT algorithms are used to address this issue by simulating quantization in a differentiable manner, allowing the network to adapt for subsequent quantization.
For instance, \cite{baskin_nice_2018, defossez_differentiable_2022} suggested injecting pseudo-quantization noise into full-precision network training.
Alternatively, \cite{esser_learned_2020, choi_pact_2018, jung_joint_2018} uses a straight-through estimator \cite{bengio_estimating_2013} to approximate the gradients of the discontinuous quantization function on the backward pass. 
\cite{gong_differentiable_2019,  qin_forward_2020, defossez_differentiable_2022} proposes a smooth approximation of this stair-step function. 
In addition, the listed base methods could be improved by using knowledge distillation~\cite{polino_model_2018}, progressive quantization~\cite{zhuang_effective_2021}, stochastic precision~\cite{dong_learning_2017}, Batch Normalization re-estimation~\cite{louizos_relaxed_2019}, additional regularization~\cite{lin_defensive_2019, alizadeh_gradient_2020}, and various non-uniform quantization extensions~\cite{yamamoto_learnable_2021, liu_nonuniform--uniform_2022, dong_hawq-v2_2020}. 

% \textbf{Research gap.} We found out that ...

\section{Background}

Quantization allows for reducing inference time and power consumption by decreasing the precision of matrix multiplication factors in convolutions or fully-connected layers.
Quantization can result in quality degradation, so \ac{qat} is applied to recover the quality and ensure model resilience.
During QAT, the original convolution operation, denoted as $X * W$, with input $X$ and weight $W$, is transformed into $Q(X) * Q(W)$, where $Q$ is a pseudo-quantization function that allows back-propagation and $*$ represents a convolutional operator.
LSQ~\cite{esser_learned_2020} is the current \ac{sota} pseudo-quantization function, so we also use it in our experiments.

When quantizing re-parameterized models, there are several options available.
\begin{enumerate}
\item Apply re-parametrization and \ac{ptq} \cite{banner_post_2018,nagel_up_nodate,cai_zeroq_2020} successively. Research works \cite{ding_reparameterizing_2023,chu_make_2022} show that re-parametrization can lead to PTQ\-/unfriendly distributions, resulting in a significant quality drop after \ac{ptq} application.
\item Apply re-parametrization and \ac{qat} successively. 
Trained re-parametrized full-precision blocks are converted into single layers, and standard QAT is applied to each of them.
% Here, trained re-parametrized full-precision blocks are converted into single layers, and then any standard QAT algorithm is applied to each of them.
\item Apply re-parametrization and \ac{qat} simultaneously as follows. Quantize each layer inside a re-parametrized block independently and then merge those layers into a single quantized layer only after QAT.
This option is impractical as merging quantized convolutions results in a convolution with a higher bit width while binary and lower bit quantization is much more challenging than four or eight-bit quantization~\cite{xia_basic_2023}. Consider a simple multiplication function $f(x) = x w$, which is re-parametrized as follows $ R(x) = x w_1 w_2$, with the resulting merged weight $w = w_1 w_2$. When $w_1$ and $w_2$ are in FP32, their multiplication usually incurs a negligible loss of precision. However, if $w_1$ and $w_2$ are quantized to 2 bits, represented by integers in the range [0, 3], their multiplication results in an integer in the range [0, 9], requiring at least 4 bits for storage. 
% several convolutional INT-8 kernels can not be re-parameterized into a single INT-8 kernel without significant precision loss. 
% For instance, merging two 4-bit sequential convolutions will result in an 8-bit convolution.

% Let us consider a modern re-parametrization block OREPA~\cite{hu_online_2022} with eight
% convolutions. To achieve an eight-bit resulting convolution, we would need to independently quantize each
% block convolution into one bit. In this way, making a four-bit quantization is
% impossible for the OREPA block. Furthermore, binary quantization is much more challenging than four or eight-bit quantization~\cite{xia_basic_2023}.

\item We introduce a novel approach where QAT and re-parametrization are applied simultaneously by performing pseudo-quantization on top of the merged re-parametrized block.
\end{enumerate}
Options 1-3 are based on well-known approaches to regular model quantization, while option 4 is our novel approach designed specifically for the quantization of re-parametrized models. 
We exclude options 1 and 3 as they are either not generic or known to produce unsatisfactory results for the re-parametrized models. In our experiments, we compare option 2 to our proposed RepQ method (option 4) to demonstrate its effectiveness.

% Such an approach is not generic because the bit widths of a resulting layer will exceed the block layers' bit widths.
% Multiplication of two 4-bit integers will be an 8-bit integer; similarly,
% merging two 4-bit sequential convolutions will result in an 8-bit convolution. This poses a
% challenge for the quantization of modern re-parametrized blocks that contain numerous convolutions
% within a single block. 
% depend on the re-parametrization block.
% For example, the OREPA~\cite{hu_online_2022} block contains eight convolutions.
% Given the minimum bit width of one for each convolution, the resulting bit width of a merged convolution will be eight-bit.
% Therefore, it is not possible to quantize a merged layer in four bits this way.
% Furthermore, binary quantization is much more challenging than four or eight-bit quantization.

\section{Proposed Methods}
\label{method}

In Section~\ref{NoBNCase}, we describe the quantization strategy of re-parameterized blocks without \ac{bn} layers and introduce a general RepQ training framework.
For blocks with \ac{bn} layers, we provide two alternative extensions in Sections~\ref{BNCase} and \ref{BNestCase}.

\subsection{RepQ: Quantization-aware training with re-parametrization}
\label{NoBNCase}

\begin{figure}
  \centering
  % \newsavebox{\myimage}% Box that will store image
  % \begin{lrbox}{\myimage}
    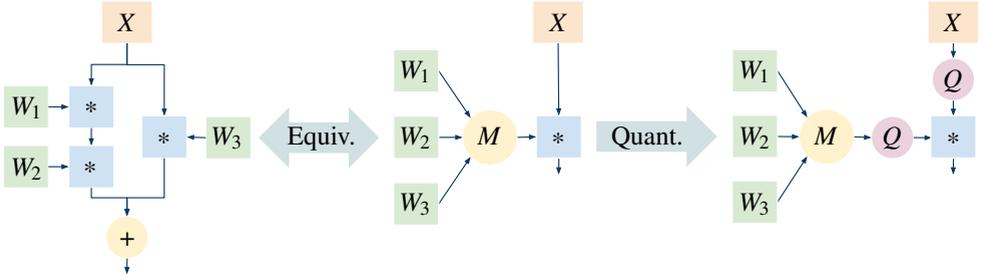% Actual image and text
  % \end{lrbox}
  % \resizebox{5in}{1.3in}{\usebox{\myimage}}% Resize box
  \caption{This picture illustrates the application of a re-parametrized quantization-aware training (RepQ) to a single layer with convolutions with input $X$ and parameters $W_1, W_2, W_3$. The left plot illustrates the re-parametrization block substituting a single convolutional layer used as an example. 
The middle plot is an equivalent transformation (in terms of the gradient flow) to the left plot. Note that the computation order is different. The right plot shows how the pseudo-quantization functions $Q$ are inserted to perform RepQ.}
  \label{noBN}
  % <caption> <label>
\end{figure}

Let us first consider the scenario in which \ac{bn} is not used in a re-parametrized block. The authors of \cite{hu_online_2022} notice that it is possible to reduce training time by merging the re-parametrized blocks without \ac{bn} into a single convolution while still optimizing the extended set of weights introduced by re-parametrization.
This section shows how this merged training benefits \ac{qat}. 

To illustrate the concept of re-parametrization, we use the simple example shown in Figure~\ref{noBN}: $ R(X, W) = X * W_1 * W_2 + X * W_3 $.
$R(X, W)$ denotes a re-parametrized block that replaces a single convolution during training. 
We can simplify this block to a single convolution with weight $M$ by deducing that $R(X, W) = X * (W_1 * W_2 + W_3) = X * M(W_1, W_2, W_3)$. In a broader sense, $M$ is a differentiable function that maps the block's trainable parameters to the weight of the final converted convolution.
This example easily generalizes to the other re-parametrization strategies, where the re-parametrized block has a form of $R(X, W) = X * M(W_1, \ldots , W_n)$ or, roughly speaking, re-parametrizations without \ac{bn}.
The articles introducing novel re-parametrized blocks regularly provide the formulas necessary to compute $M$, so we do not repeat them here. 

Notably, merged training does not affect the gradient flow, as the two sides of the equation are numerically equivalent during both forward and backward passes. However, the merged weight $M$ is explicitly computed on the right side.  

Now it is easy to see that the pseudo-quantization function can be applied on top of $M$:
\begin{equation}
    X * M(W_1, \ldots, W_n) \to Q(X) * Q(M(W_1, \ldots, W_n)).
\end{equation}
As a result, function $Q(M(W_1, \ldots, W_n))$ will equal the quantized weight used on the inference.
Since $Q$ and $M$ are differentiable functions, the gradient propagates smoothly to the weights $W_1, \ldots, W_n$. 
The combination of re-parametrization with introducing $M$ and pseudo-quantization function $Q$ constitutes the RepQ approach and enables end-to-end quantization-aware training. This is the main algorithm used in all RepQ experiments.

\subsection{RepQ-BN: Merging batch normalization}
\label{BNCase}

Many \ac{sota} re-parametrizations use batch normalization. 
Several papers show that BN is an essential component of their blocks. BN's removal leads to a significant performance drop. 
Since we aim to provide a quantization strategy that generalizes well to diverse re-parametrizations, we study how to handle \ac{bn} in \ac{qat}. 

The first option is fusing the \ac{bn} with the preceding convolution during training, described in this section. A similar procedure was proposed in ~\cite{jacob_quantization_2018, jin2021f8net} to achieve integer\-/arithmetic-only quantization. In our case, folding \ac{bn} reduces the task to the no-BN case described in the previous section, \ref{NoBNCase}.

\newcommand{\bigbracket}[1]{\Bigl [ #1 \Bigr ]}
\begin{equation}
\label{BNMergeFormula}
\begin{gathered}
     \mathcal{BN}(X * W) = \frac{X * W  - \mathbb{E} \bigbracket{X * W}}{\sqrt{\mathbb{V}\bigbracket{X * W} + \epsilon}} \gamma + \beta = \\
      = X * \frac{W}{ \sqrt{\mathbb{V} \bigbracket{X * W}  + \epsilon}} \gamma - \frac{\mathbb{E} \bigbracket{X * W}}{\sqrt{\mathbb{V}\bigbracket{X * W}  + \epsilon}} \gamma + \beta = X * M(X, W) + b(X, W).
\end{gathered}
\end{equation}

 %      \    /\
 %       )  ( ')
 %      (  /  )
 % mau   \(__)|

The equation \eqref{BNMergeFormula} shows that a convolution followed by \ac{bn} is equivalent to a single convolutional operator. 
However, its parameters are dependent on the input $X$ through the batch statistics, mean and variance. 
Formally, re-parametrization with \ac{bn} has the form of $R(X, W) = X * M(X, W_1, \ldots, W_n)$. 
Algorithm~\ref{Algo:BNfuse} shows how to compute $M$ and apply quantization in practice for a simple case of $R(X, W) = \mathcal{BN}(X * W)$. By fusing \ac{bn} with preceding convolutions, we reduce the task of merging weights to the no-BN case described in Subsection~\ref{NoBNCase}.
We call this variant of our approach RepQ-BN.

\algrenewcommand\algorithmicrequire{\textbf{Given:}}

\begin{algorithm}
\caption{Fusing BN with Convolution for Quantization}\label{alg:cap}

\begin{algorithmic}[1]
\Require $R(X, W) = \mathcal{BN}(X * W)$ \Comment{a simple re-parametrization of BN after a convolution}
% \Result a
\State $Y = X * W$

\State $\mu, V = \mathbb{E} \bigbracket{Y}, \mathbb{V} \bigbracket{Y}$ \Comment{computing BatchNorm statistics}

\State $\hat{\mu} = (1 - m) \cdot \hat{\mu} + m \cdot \mu$ \Comment{updating cumulative moving mean ($m$ denotes momentum)}

\State $\hat{V} = (1 - m) \cdot \hat{V} + m \cdot V$ \Comment{updating cumulative moving variance}

\State $M(X, W) = \frac{W}{ \sqrt{V  + \epsilon} }\cdot \gamma$ \Comment{computing merged weight}
\State $ b(X, W) = - \frac{\mu}{ \sqrt{V  + \epsilon}} \cdot \gamma + \beta $ \Comment{computing merged bias }
\State $R_{q}(X, W) = Q(X) * Q(M(X, W)) + b(X, W)$ \Comment{quantized re-parametrized layer}
\end{algorithmic}
\label{Algo:BNfuse}
\end{algorithm}

\subsection{RepQ-BNEst: Batch normalization estimation}
\label{BNestCase}
An observant reader may have noticed that in Algorithm~\ref{Algo:BNfuse}, the convolutional operator is computed twice, first in line 1 and then again in line 7. While the additional computation in line 1 is used to calculate the \ac{bn} statistics $\mu$ and $V,$ the question arises whether it is necessary to perform such a computationally expensive convolution to determine the mean and variance of the output. Here we propose a novel method of estimating \ac{bn} running statistics based on inputs and weights without computing the convolution.

For simplicity, let us consider the case of a $1 \times 1$ convolution. The $1 \times 1$ convolution can be viewed as a matrix multiplication of the input $X$ of the shape $[B \cdot H \cdot D, IN]$ and weight $W$ of shape $[IN, OUT]$, where $B$ is the batch size, $H$ is the image height, $D$ is the image width, $IN$ is the number of input channels, and $OUT$ is the number of output channels. Consider computing \ac{bn} mean statistics,

\begin{equation}
    \mathbb{E} \left[  X W \right] = \mathbb{E} \left[ X \right] W.
\end{equation}

% \ik{Variables B  H D IN OUT should be in italics}
The equation suggests we can first calculate the per-channel mean over the batch, height, and width dimensions and then multiply the result by the weight matrix. This approach has a computational complexity of $\text{O}(B \cdot H \cdot D \cdot IN)$, making it more favourable than the naive solution, which has a complexity of $\text{O}(B \cdot H \cdot D \cdot IN \cdot OUT)$. Additionally, \ac{bn} estimation allows us to avoid storing the feature map $XW$ on the GPU.

For exact variance computation, a similar reduction is not possible due to the need to calculate the input covariance matrix. As a solution, we propose to approximate the covariance matrix with a diagonal form,

\begin{equation}
   \mathbb{V} \bigbracket{X W} = 
   % \mathcal{D} \bigbracket{\mathrm{Cov}(XW, XW)}=
   \mathcal{D} \bigbracket{ W^T  \mathrm{Cov}(X, X)  W 
    }
 \approx  \mathcal{D} \bigbracket{ W^T  \mathcal{D} \bigbracket{\mathrm{Cov}(X, X)}  W } =
 \mathbb{V}\bigbracket{X} W^2, 
\end{equation}
where $W^2$ is an element-wise square of $W$, $\mathrm{Cov}$ is the sample covariance matrix, and $\mathcal{D}$ is the diagonalizing operator that leaves only diagonal matrix elements non-zero.
As a result of \ac{bn} estimation, the variance is substituted with another quadratic statistic of the weight and the output that estimates variance but is computationally more efficient for quantization-aware re-parametrization.
The above formulas generalize to arbitrary weight shapes as follows (for more details, we refer to supplementary materials),
\begin{equation}
    \mathbb{E} \bigbracket{X * W} \approx \tilde{\mathbb{E}} \bigbracket{X * W} = \mathbb{E} \bigbracket{X} \cdot \sum_{h, d} W_{h, d} ,
\end{equation}
\begin{equation}
   \mathbb{V} \bigbracket{X * W} \approx \tilde{\mathbb{V}} \bigbracket{X * W} =
   \mathbb{V}\bigbracket{X} \cdot \sum_{h, d} W_{h, d}^2.
   % \mathcal{D} \bigbracket{  \sqrt{\sum_{h, d} W_{h,d}^2}^T \cdot \mathcal{D} \bigbracket{\mathrm{Cov}(X, X)} \cdot \sqrt{\sum_{h, d} W_{h, d}^2} }. 
\end{equation}

To sum up, for \ac{bn} Estimation, we modify Algorithm~\ref{Algo:BNfuse} by replacing the calculation of mean $\mu$ and variance $V$ with $\tilde{\mathbb{E}}$ and $\tilde{\mathbb{V}}$, respectively,  in line 2. In addition, we skip computing $Y$ in line 1.
We call this variant of our approach RepQ-BNEst.

\section{Experiments}

\subsection{Experimental setup}
% \input{tables/hyperparams}
% refer to particular articles that uses these hyperparameters
% we conduct experiments using architectures and datasets from articles on reparametrization

\textbf{Architectures.}
We evaluate the performance of RepQ on three architectures: ResNet\-/18, VGG, and ECBSR~\cite{zhang_edge-oriented_2021}.
For ResNet-18, we employ two re-parametrization techniques, the well-known ACNets \cite{ding_acnet_2019} and the recently published OREPA~\cite{hu_online_2022}, which we refer to as AC-ResNet-18 and OREPA-ResNet-18, respectively.
For VGG, we use the \ac{sota} RepVGG \cite{ding_repvgg_2021} approach to re-parametrization with two network depth variants, RepVGG-A0 and RepVGG-B0.
The ECBSR is a re-parametrization approach developed for the super\-/resolution problem. Our focus on evaluating RepQ with lightweight versions of SOTA architectures was motivated by their suitability for mobile devices.

\textbf{Comparisons.}
We compare our RepVGG quantization results with QARepVGG, a quantization-friendly version of VGG introduced in ~\cite{chu_make_2022}. To the best of our knowledge, no studies provide quantization results for re-parametrized architectures apart from RepVGG. To further support the main claims of our article, we have designed baselines, which are described in the following paragraph.

\textbf{Baselines.} Quantized model training includes two sequential stages: (1) a regular full-precision (FP) pre-training and (2) \ac{qat}.
Weights pre-trained on the FP stage are used to initialize the quantized model at the beginning of the second stage. 
% This initialization is important for quantized model performance.
We compare our re-parametrized quantization-aware training RepQ that uses re-parametrization during QAT with the baselines that do not use re-parametrization in the QAT stage.
In particular, we compare RepQ to plain and merged baselines schematically described in Table~\ref{table:abbreviations}:
\begin{itemize}
    \item \textbf{Plain.} Regularly trained model with no re-parametrization in both stages. 
    \item \textbf{Merged.} The re-parametrized model is trained in the FP stage. Re-parametrized blocks are merged back into single convolutions, and the merged weights are used to initialize the quantized model. Such initialization can be profitable for subsequent quantization because re-parametrized models usually have better metrics than regular models.
    % \item \textbf{RepQ.} Variations of the proposed method described in \ref{method}.
\end{itemize}
\begin{table}[H]
\begin{center}
\begin{tabular}{c|cc|cc}
 % \centering
% \hline 
\multirow{2}{*}{} & \multicolumn{2}{c|}{FP pre-training stage} & \multicolumn{2}{c}{QAT stage} \\
                  & Regular           & Re-parametrized        & Regular     & Re-parametrized \\ \hline
Plain             & \checkmark        &                        & \checkmark  &                 \\
Merged            &                   & \checkmark             & \checkmark  &                 \\
RepQ              &                   & \checkmark             &             & \checkmark   
\end{tabular}
\end{center}
\caption{Quantization strategies: baselines (Plain, Merged) and our proposed method(RepQ).}
\label{table:abbreviations}
\end{table}

% We refer to our method as RepQ, RepQ-BN, and RepQ-BNEst, depending on the BN handling algorithm. 
% As the ECBSR block initially does not contain BN, we do not use BN for quantization either. 
% For AC-Nets and RepVGG blocks containing BN, we test two quantization alternatives: RepQ-BN \ref{BNCase} and RepQ-BNEst \ref{BNestCase}. 
% In all quantization experiments, we use state-of-the-art pseudo-quantization functions $Q$, proposed in the LSQ paper~\cite{esser_learned_2020}, as it provides superior results to other quantization approaches. 

 % добавить референс на аппендикс

\textbf{Training pipeline.}
Generally, re-parametrization changes only the model architecture.
For example, imagine that all convolutions are replaced by the following re-parametrization blocks $R(X) = BN(X * W) + X$.
The simplified pseudo-code of this block is shown below.
% We provide a simplified pseudo code for it below.
% \scalebox{0.68}{
% \vspace{-0.1cm}
\begin{lstlisting}[language=Python,basicstyle=\scriptsize,]
class ReparametrizedBlock:
  def kernel(x, training):
    # Proposed differentiable weight function, M.
    BN_stats = self.BN_stats(x, self.conv.weight, training)
    weight = fuse_BN(self.conv.weight, BN_stats)
    return fuse_residual(weight)

  def BN_stats(x, weight, training): 
    if self.BNEst:
      return BNEst(x, weight, training) # Use Eq. 5-6.
    else: 
      return  BN(x, weight, training) # Use Algorithm 1..

  def call(x, training=False):
    # "training" defines the training or testing mode.
    if self.quantize:
      return conv2d(Q(x), Q(self.kernel(x, training))) # Quantized training.
    elif self.BNEst:
      return conv2d(x, self.kernel(x, training)) # BN estimation for FP training.
    else:
      return self.bn(self.conv(x)) + x   # Regular FP training with BN.
\end{lstlisting}
% \vspace{-0.1cm}
% }
Once full-precision training converges, the "self.quantize" parameters are set to True, and training is repeated with minor changes to hyperparameters, described in the supplementary.

\textbf{Implementation details.} 
% We trained each full-precision model according to the setup described in each 
We trained full-precision models in accordance with the re\-/parametrization articles' setup using official repositories. That is why the baseline for the same model architecture may slightly differ for various re-parametrization blocks. 
We trained quantized models with the same hyperparameter setup as the full-precision models, except for learning rate adjustments.
All quantized models are initialized with corresponding pre-trained full-precision weights; quantization steps are initialized using MinError~\cite{bhalgat_lsq_nodate} initialization for the first batch. 
For reproducibility, we provide hyperparameters in 
% Table~\ref{table:ParamsSetup}. 
the supplementary materials.

\subsection{Experimental results}
\label{sec:results}

We present our main results in Tables~\ref{table:classification} and \ref{table:ecbsr}.
They demonstrate that the proposed variants of RepQ consistently outperform the baselines of all the tested architectures and re-parametrized blocks.
Table \ref{table:ecbsr} exhibits the benefits of RepQ on a wide range of super\-/resolution datasets.
For all the studied classification models, the 8-bit RepQ performance exceeds the full-precision result, while the QARepVGG~\cite{chu_make_2022} exhibits a quality drop. The gap between our RepQ and QARepVGG exceeds 1\% for both RepVGG-A0 and RepVGG-B0.
For RepVGG and ResNet, the best result is achieved either by RepQ-BN or RepQ-BNEst. In particular, RepQ-BN is slightly better than RepQ-BNEst in eight-bit experiments; nevertheless, RepQ-BNEst achieves better results in four-bit quantization.
Surprisingly, RepQ-BNEst outperforms RepQ-BN by a considerable margin on the 4-bit RepVGG-A0 and RepVGG-B0. 
In addition, we measured the ResNet-18 training speed on 2 V100 GPUs.
RepQ-BNEst allows for a \textbf{25\%} training time reduction compared to RepQ-BN. 
The merged baseline demonstrates inconsistency on different studied architectures. For RepVGG-B0, the merged results are quite close to RepQ; however, for Resnet-18 and ECBSR, it is inferior to plain models.

\begin{table}[H]
\begin{center}
\begin{tabular}{ccccc}
 % \centering
% \hline 
\multirow{2}{*}{Model} & 
\multirow{2}{*}{Method} &
\multicolumn{3}{c}{Precision } \\
 & & \makecell{32 (FP) } & 8 & 4  \\
\hline 

\multirow{5}{*}{RepVGG-A0~\cite{ding_repvgg_2021} } 
&QARepVGG~\cite{chu_make_2022} & \underline{72.40} & 71.90 & - \\
& Plain & 70.91 & 71.52  & 69.90 \\
&  Merged & \text { - } & 72.49 & \text{69.21} \\
& RepQ-BN & 72.25 & \textbf{73.11} &  \underline{70.31} \\
&  RepQ-BNEst  & \textbf{72.43} & \underline{72.94}  & \textbf{71.12} \\

\hline \multirow{5}{*}{ RepVGG-B0~\cite{ding_repvgg_2021} } 
& QARepVGG~\cite{chu_make_2022} & 75.10             & 74.60 & - \\
&  Plain                        & 73.48             & 74.79  & 72.52 \\
&  Merged                       & \text { - }       & 75.48 &  \text { 73.06 } \\
&  RepQ-BN                      & \underline{75.27} & \textbf{75.60} & \underline{73.51} \\ 
& \text { RepQ-BNEst }          & \textbf{75.38}    & \underline{75.53}          & \textbf{ 74.01 } \\
\hline

\multirow{4}{*}{AC-Resnet-18~\cite{ding_acnet_2019}} 
& Plain & 70.24 &	70.77 & 70.00  \\
& Merged &  -  & 70.64 & 68.25  \\
& RepQ-BN & \textbf{71.01} &	\textbf{71.40}	 & \underline{70.07}  \\ 
& RepQ-BNEst & \underline{70.80} & \underline{ 71.23 } & \textbf{70.20}  \\
\hline

\multirow{3}{*}{OREPA-Resnet-18~\cite{hu_online_2022}}
& Plain  & \underline{71.24} &	\underline{71.76} & \textbf{71.49} \\
& Merged & -  & 69.58 & 69.96 \\
& RepQ & \textbf{72.07} &	\textbf{72.32}	& \textbf{71.49} \\ 
\hline
% \caption{Some caption here}
\end{tabular}
\end{center}
\caption{ImageNet~\cite{russakovsky_imagenet_2014} top-1 accuracy for different quantization strategies for re-parametrized~NNs.
The best result is emphasized with bold text, and the second best is underscored.
}
\label{table:classification}
\end{table}
\begin{table}[H] %t
\begin{center}

\begin{tabular}{cccccc}
% \vskip -0.3in
Precision &  Method & Set5~\cite{bevilacqua_low-complexity_2012} & Set14~\cite{zeyde_single_2010} &
B100~\cite{martin_database_2001} & 
U100~\cite{huang_single_2015} \\
\hline 
32 & & 36.90 & 32.62 & 31.43 & 29.66 \\
\hline 
\multirow{3}{*}{8} 
& Plain & 36.81 & 32.55 & 31.39 & 29.57 \\
& Merged & 36.86 & 32.54 & 31.39 & 29.53 \\
& RepQ  & \textbf{36.91} & \textbf{32.57} & \textbf{31.40}  & \textbf{29.60}  \\

\hline
\multirow{3}{*}{4} 
& Plain & 36.20 & 32.16 & \textbf{31.09} & 28.91 \\
& Merged & 36.16 & 32.15 & 31.08 & \textbf{28.95} \\
& RepQ & \textbf{36.30} & \textbf{32.19} & \textbf{31.09} & \textbf{28.95} \\
 
\hline 
\multirow{3}{*}{2} 
& Plain & 34.82  & 31.30 & 30.46 & 27.92 \\
& Merged & 34.74 & 31.28 & 30.44 & 27.91 \\
& RepQ & \textbf{34.89} & \textbf{31.36} & \textbf{30.50} & \textbf{28.00} \\
\hline
\end{tabular}
\end{center}
\caption{PSNR metric for ECBSR-m4c8~\cite{zhang_edge-oriented_2021} architecture trained on DIV2K~\cite{agustsson_ntire_2017} dataset, and x2 scaling. }
\label{table:ecbsr}
\end{table}

\subsection{Discussion}

Our results demonstrate that 8-bit quantization can improve the quality of re-parametrized models.
A similar behaviour is also observed in plain models~\cite{zhou_dorefa-net_2016,esser_learned_2020,solodskikh_towards_2022} and usually explained by an additional regularization effect of quantization.
Four-bit quantization with the RepQ method results in a minor quality drop.
At the same time, the number of bit-operations is reduced by four times compared to 8-bit models.
Interestingly, 4-bit RepVGG-B0 has two times less bit-operations than 8-bit RepVGG-A0, while its accuracy is higher.
This makes 4-bit RepVGG-B0 more favourable for deployment.

\textbf{Limitations.}
The main limitation of re-parametrization and RepQ is the increase in training time (TT). Let us take as an example ResNet-18 re-parametrized with two different blocks: ACNets and OREPA. The table provides the relative training time of the re-parametrized models for Plain FP and QAT networks. The TT increase introduced by RepQ is comparable to the one introduced by re\-/parametrization on full-precision training for both ACNets and OREPA blocks. Although RepQ-BN experiences a longer training due to the twice-forward computation, RepQ-BNEst mitigates this issue.
Overall, despite the training time overhead, re-parametrization and RepQ is beneficial when the trade-off between inference time and model quality is the priority.

% \scalebox{0.68}{
\begin{table}[H] 
\begin{center}
\begin{tabular}{c|cc|cccc}
 \centering
\multirow{2}{*}{Block} & 
\multicolumn{2}{c}{FP} 
& \multicolumn{4}{c}{QAT} \\
& Plain  & Re-parametrized & Plain     & RepQ & RepQ-BN & RepQ-BNEst \\ \hline
ACNets & 100\% & 200\% & 100\% & - & 195\% & 150\% \\
OREPA & 100\% & 210\% & 100\% & 165\% & - & - \\ 
\end{tabular}
\caption{Training time overhead for different re-parametrization strategies on ResNet-18.}
\label{table:time}
\end{center}
\end{table}
% }

\section{Conclusions}

This paper introduces RepQ, a \ac{qat} strategy specifically designed for re-parametrized models. During training, RepQ merges re-parametrized blocks into a single convolution and applies a pseudo-quantization function on top of the merged weight. To be able to quantize arbitrary re-parametrization blocks, we provide a natural solution for merging non-linear batch normalization layers inside the blocks. Furthermore, we enhance this solution by estimating \ac{bn} statistics and thus achieve a speed-up. We conduct an extensive experimental evaluation for diverse re-parametrized blocks and model architectures. Results show that RepQ surpasses existing solutions and provides lossless 8-bit quantization for the re-parametrized models. Overall, RepQ expands the applicability of re-parametrization to the field of quantized \ac{nn} with an easy-to-implement approach.

% \ik{please check the references below: reference 11/19 has no source, Reference 12 is strange (In <peoples' names>), Reference 14 no year}
% \printbibliography
\bibliography{egbib}

\begin{thebibliography}{57}
\providecommand{\natexlab}[1]{#1}
\providecommand{\url}[1]{\texttt{#1}}
\expandafter\ifx\csname urlstyle\endcsname\relax
  \providecommand{\doi}[1]{doi: #1}\else
  \providecommand{\doi}{doi: \begingroup \urlstyle{rm}\Url}\fi

\bibitem[Agustsson and Timofte(2017)]{agustsson_ntire_2017}
Eirikur Agustsson and Radu Timofte.
\newblock {NTIRE} 2017 {Challenge} on {Single} {Image} {Super}-{Resolution}:
  {Dataset} and {Study}.
\newblock In \emph{Proceedings of the {IEEE}/{CVF} {Conference} on {Computer}
  {Vision} and {Pattern} {Recognition} {Workshops}}, 2017.

\bibitem[Alizadeh et~al.(2020)Alizadeh, Behboodi, van Baalen, Louizos,
  Blankevoort, and Welling]{alizadeh_gradient_2020}
Milad Alizadeh, Arash Behboodi, Mart van Baalen, Christos Louizos, Tijmen
  Blankevoort, and Max Welling.
\newblock Gradient l1 {Regularization} for {Quantization} {Robustness}.
\newblock In \emph{Proceedings of the {International} {Conference} on
  {Learning} {Representations}}, 2020.

\bibitem[Arora et~al.(2018)Arora, Cohen, and Hazan]{arora_optimization_2018}
Sanjeev Arora, Nadav Cohen, and Elad Hazan.
\newblock On the {Optimization} of {Deep} {Networks}: {Implicit} {Acceleration}
  by {Overparameterization}.
\newblock In \emph{Proceedings of {The} {International} {Conference} on
  {Machine} {Learning}}, 2018.

\bibitem[Banner et~al.(2018)Banner, Nahshan, and Soudry]{banner_post_2018}
Ron Banner, Yury Nahshan, and Daniel Soudry.
\newblock Post training 4-bit quantization of convolutional networks for
  rapid-deployment.
\newblock In \emph{Proceedings of the {Conference} on {Neural} {Information}
  {Processing} {Systems}}, 2018.

\bibitem[Baskin et~al.(2018)Baskin, Liss, Chai, Zheltonozhskii, Schwartz,
  Giryes, Mendelson, and Bronstein]{baskin_nice_2018}
Chaim Baskin, Natan Liss, Yoav Chai, Evgenii Zheltonozhskii, Eli Schwartz, Raja
  Giryes, Avi Mendelson, and Alexander~M. Bronstein.
\newblock {NICE}: {Noise} {Injection} and {Clamping} {Estimation} for {Neural}
  {Network} {Quantization}.
\newblock In \emph{Proceedings of the {International} {Conference} on
  {Learning} {Representations}}, 2018.

\bibitem[Bengio et~al.(2013)Bengio, Léonard, and
  Courville]{bengio_estimating_2013}
Yoshua Bengio, Nicholas Léonard, and Aaron~C. Courville.
\newblock Estimating or {Propagating} {Gradients} {Through} {Stochastic}
  {Neurons} for {Conditional} {Computation}.
\newblock \emph{arXiv:1308.3432}, 2013.

\bibitem[Bevilacqua et~al.(2012)Bevilacqua, Roumy, Guillemot, and
  Morel]{bevilacqua_low-complexity_2012}
Marco Bevilacqua, Aline Roumy, Christine Guillemot, and Marie-line~Alberi
  Morel.
\newblock Low-{Complexity} {Single}-{Image} {Super}-{Resolution} based on
  {Nonnegative} {Neighbor} {Embedding}.
\newblock In \emph{Proceedings of the {British} {Machine} {Vision}
  {Conference}}, 2012.

\bibitem[Bhalgat et~al.(2020)Bhalgat, Lee, Nagel, Blankevoort, and
  Kwak]{bhalgat_lsq_nodate}
Yash Bhalgat, Jinwon Lee, Markus Nagel, Tijmen Blankevoort, and Nojun Kwak.
\newblock {LSQ}+: {Improving} low-bit quantization through learnable offsets
  and better initialization.
\newblock In \emph{Camera-ready for {Joint} {Workshop} on {Efficient} {Deep}
  {Learning} in {Computer} {Vision}, {CVPR}}, 2020.

\bibitem[Cai et~al.(2020)Cai, Yao, Dong, Gholami, Mahoney, and
  Keutzer]{cai_zeroq_2020}
Yaohui Cai, Zhewei Yao, Zhen Dong, Amir Gholami, Michael~W Mahoney, and Kurt
  Keutzer.
\newblock Zeroq: {A} novel zero shot quantization framework.
\newblock In \emph{Proceedings of the {IEEE}/{CVF} {Conference} on {Computer}
  {Vision} and {Pattern} {Recognition}}, 2020.

\bibitem[Choi et~al.(2018)Choi, Wang, Venkataramani, Chuang, Srinivasan, and
  Gopalakrishnan]{choi_pact_2018}
Jungwook Choi, Zhuo Wang, Swagath Venkataramani, Pierce I-Jen Chuang,
  Vijayalakshmi Srinivasan, and Kailash Gopalakrishnan.
\newblock Pact: {Parameterized} clipping activation for quantized neural
  networks.
\newblock \emph{arXiv:1805.06085}, 2018.

\bibitem[Chu et~al.(2022)Chu, Li, and Zhang]{chu_make_2022}
Xiangxiang Chu, Liang Li, and Bo~Zhang.
\newblock Make {RepVGG} {Greater} {Again}: {A} {Quantization}\-/aware
  {Approach}.
\newblock \emph{arXiv:2212.01593}, 2022.

\bibitem[Denil et~al.(2013)Denil, Shakibi, Dinh, Ranzato, and
  Freitas]{denil_predicting_2013}
Misha Denil, Babak Shakibi, Laurent Dinh, Marc'Aurelio Ranzato, and Nando~de
  Freitas.
\newblock Predicting {Parameters} in {Deep} {Learning}.
\newblock In \emph{Proceedings of the {Conference} on {Neural} {Information}
  {Processing} {Systems}}, 2013.

\bibitem[Ding et~al.(2019)Ding, Guo, Ding, and Han]{ding_acnet_2019}
Xiaohan Ding, Yuchen Guo, Guiguang Ding, and Jungong Han.
\newblock {ACNet}: {Strengthening} the {Kernel} {Skeletons} for {Powerful}
  {CNN} via {Asymmetric} {Convolution} {Blocks}.
\newblock In \emph{Proceedings of the {IEEE}/{CVF} {International} {Conference}
  on {Computer} {Vision}}, 2019.

\bibitem[Ding et~al.(2021{\natexlab{a}})Ding, Hao, Tan, Liu, Han, Guo, and
  Ding]{ding_resrep_2021}
Xiaohan Ding, Tianxiang Hao, Jianchao Tan, Ji~Liu, Jungong Han, Yuchen Guo, and
  Guiguang Ding.
\newblock {ResRep}: {Lossless} {CNN} {Pruning} via {Decoupling} {Remembering}
  and {Forgetting}.
\newblock In \emph{Proceedings of the {IEEE}/{CVF} {International} {Conference}
  on {Computer} {Vision}}, 2021{\natexlab{a}}.

\bibitem[Ding et~al.(2021{\natexlab{b}})Ding, Zhang, Han, and
  Ding]{ding_diverse_2021}
Xiaohan Ding, Xiangyu Zhang, Jungong Han, and Guiguang Ding.
\newblock Diverse {Branch} {Block}: {Building} a {Convolution} as an
  {Inception}-like {Unit}.
\newblock In \emph{Proceedings of the {IEEE}/{CVF} {Conference} on {Computer}
  {Vision} and {Pattern} {Recognition}}, 2021{\natexlab{b}}.

\bibitem[Ding et~al.(2021{\natexlab{c}})Ding, Zhang, Ma, Han, Ding, and
  Sun]{ding_repvgg_2021}
Xiaohan Ding, Xiangyu Zhang, Ningning Ma, Jungong Han, Guiguang Ding, and Jian
  Sun.
\newblock Repvgg: {Making} vgg-style convnets great again.
\newblock In \emph{Proceedings of the {IEEE}/{CVF} {Conference} on {Computer}
  {Vision} and {Pattern} {Recognition}}, 2021{\natexlab{c}}.

\bibitem[Ding et~al.(2023)Ding, Chen, Zhang, Huang, Han, and
  Ding]{ding_reparameterizing_2023}
Xiaohan Ding, Honghao Chen, Xiangyu Zhang, Kaiqi Huang, Jungong Han, and
  Guiguang Ding.
\newblock Re-parameterizing {Your} {Optimizers} rather than {Architectures}.
\newblock In \emph{Proceedings of the {International} {Conference} on
  {Learning} {Representations}}, 2023.

\bibitem[Dong et~al.(2017)Dong, Ni, Li, Chen, Zhu, and Su]{dong_learning_2017}
Yinpeng Dong, Renkun Ni, Jianguo Li, Yurong Chen, Jun Zhu, and Hang Su.
\newblock Learning accurate low-bit deep neural networks with stochastic
  quantization.
\newblock In \emph{Proceedings of the {British} {Machine} {Vision}
  {Conference}}, 2017.

\bibitem[Dong et~al.(2020)Dong, Yao, Arfeen, Gholami, Mahoney, and
  Keutzer]{dong_hawq-v2_2020}
Zhen Dong, Zhewei Yao, Daiyaan Arfeen, Amir Gholami, Michael~W Mahoney, and
  Kurt Keutzer.
\newblock Hawq-v2: {Hessian} aware trace-weighted quantization of neural
  networks.
\newblock In \emph{Proceedings of the {Conference} on {Neural} {Information}
  {Processing} {Systems}}, 2020.

\bibitem[Défossez et~al.(2022)Défossez, Adi, and
  Synnaeve]{defossez_differentiable_2022}
Alexandre Défossez, Yossi Adi, and Gabriel Synnaeve.
\newblock Differentiable {Model} {Compression} via {Pseudo} {Quantization}
  {Noise}.
\newblock \emph{Transactions on Machine Learning Research}, 2022.

\bibitem[Esser et~al.(2020)Esser, McKinstry, Bablani, Appuswamy, and
  Modha]{esser_learned_2020}
Steven~K. Esser, Jeffrey~L. McKinstry, Deepika Bablani, Rathinakumar Appuswamy,
  and Dharmendra~S. Modha.
\newblock {LEARNED} {STEP} {SIZE} {QUANTIZATION}.
\newblock In \emph{Proceedings of the {International} {Conference} on
  {Learning} {Representations}}, 2020.

\bibitem[Gong et~al.(2019)Gong, Liu, Jiang, Li, Hu, Lin, Yu, and
  Yan]{gong_differentiable_2019}
Ruihao Gong, Xianglong Liu, Shenghu Jiang, Tianxiang Li, Peng Hu, Jiazhen Lin,
  Fengwei Yu, and Junjie Yan.
\newblock Differentiable soft quantization: {Bridging} full-precision and
  low-bit neural networks.
\newblock In \emph{Proceedings of the {IEEE}/{CVF} {International} {Conference}
  on {Computer} {Vision}}, 2019.

\bibitem[Guo et~al.(2020)Guo, Alvarez, and Salzmann]{guo_expandnets_2020}
Shuxuan Guo, Jose~M. Alvarez, and Mathieu Salzmann.
\newblock {ExpandNets}: {Linear} {Over}-parameterization to {Train} {Compact}
  {Convolutional} {Networks}.
\newblock In H.~Larochelle, M.~Ranzato, R.~Hadsell, M.~F. Balcan, and H.~Lin,
  editors, \emph{Proceedings of the {Conference} on {Neural} {Information}
  {Processing} {Systems}}, 2020.

\bibitem[Han et~al.(2015)Han, Pool, Tran, and Dally]{han_learning_2015}
Song Han, Jeff Pool, John Tran, and William~J. Dally.
\newblock Learning both {Weights} and {Connections} for {Efficient} {Neural}
  {Network}.
\newblock In \emph{Proceedings of the {Conference} on {Neural} {Information}
  {Processing} {Systems}}, 2015.

\bibitem[He et~al.(2015)He, Zhang, Ren, and Sun]{he_deep_nodate}
Kaiming He, X.~Zhang, Shaoqing Ren, and Jian Sun.
\newblock Deep {Residual} {Learning} for {Image} {Recognition}.
\newblock In \emph{Proceedings of the {IEEE}/{CVF} {Conference} on {Computer}
  {Vision} and {Pattern} {Recognition}}, 2015.

\bibitem[Hinton et~al.(2015)Hinton, Vinyals, and Dean]{hinton_distilling_2015}
Geoffrey Hinton, Oriol Vinyals, and Jeffrey Dean.
\newblock Distilling the {Knowledge} in a {Neural} {Network}.
\newblock In \emph{{NeurIPS} {Deep} {Learning} and {Representation} {Learning}
  {Workshop}}, 2015.

\bibitem[Hu et~al.(2022)Hu, Feng, Hua, Lai, Huang, Hua, and
  Gong]{hu_online_2022}
Mu~Hu, Junyi Feng, Jiashen Hua, Baisheng Lai, Jianqiang Huang, Xiansheng Hua,
  and Xiaojin Gong.
\newblock Online {Convolutional} {Re}-parameterization.
\newblock In \emph{Proceedings of the {IEEE}/{CVF} {Conference} on {Computer}
  {Vision} and {Pattern} {Recognition}}, 2022.

\bibitem[Huang et~al.(2015)Huang, Singh, and Ahuja]{huang_single_2015}
Jia-Bin Huang, Abhishek Singh, and Narendra Ahuja.
\newblock Single image super-resolution from transformed self-exemplars.
\newblock In \emph{Proceedings of the {IEEE}/{CVF} {Conference} on {Computer}
  {Vision} and {Pattern} {Recognition}}, 2015.

\bibitem[Ioffe and Szegedy(2015)]{ioffe_batch_2015}
Sergey Ioffe and Christian Szegedy.
\newblock Batch {Normalization}: {Accelerating} {Deep} {Network} {Training} by
  {Reducing} {Internal} {Covariate} {Shift}.
\newblock In \emph{Proceedings of {The} {International} {Conference} on
  {Machine} {Learning}}, 2015.

\bibitem[Jacob et~al.(2018)Jacob, Kligys, Chen, Zhu, Tang, Howard, Adam, and
  Kalenichenko]{jacob_quantization_2018}
Benoit Jacob, Skirmantas Kligys, Bo~Chen, Menglong Zhu, Matthew Tang, Andrew
  Howard, Hartwig Adam, and Dmitry Kalenichenko.
\newblock Quantization and training of neural networks for efficient
  integer-arithmetic-only inference.
\newblock In \emph{Proceedings of the {IEEE}/{CVF} {Conference} on {Computer}
  {Vision} and {Pattern} {Recognition}}, 2018.

\bibitem[Jaderberg et~al.(2014)Jaderberg, Vedaldi, and
  Zisserman]{jaderberg_speeding_2014}
Max Jaderberg, Andrea Vedaldi, and Andrew Zisserman.
\newblock Speeding up {Convolutional} {Neural} {Networks} with {Low} {Rank}
  {Expansions}.
\newblock In \emph{Proceedings of the {British} {Machine} {Vision}
  {Conference}}, 2014.

\bibitem[Jin et~al.(2021)Jin, Ren, Zhuang, Hanumante, Li, Chen, Wang, Yang, and
  Tulyakov]{jin2021f8net}
Qing Jin, Jian Ren, Richard Zhuang, Sumant Hanumante, Zhengang Li, Zhiyu Chen,
  Yanzhi Wang, Kaiyuan Yang, and Sergey Tulyakov.
\newblock F8net: {Fixed-Point} 8-bit {Only} {Multiplication} for {Network}
  {Quantization}.
\newblock 2021.

\bibitem[Jung et~al.(2018)Jung, Son, Lee, Son, Kwak, Han, and
  Choi]{jung_joint_2018}
Sangil Jung, Changyong Son, Seohyung Lee, Jinwoo Son, Youngjun Kwak, Jae-Joon
  Han, and Changkyu Choi.
\newblock Joint training of low-precision neural network with quantization
  interval parameters.
\newblock \emph{arXiv:1808.05779}, 2018.

\bibitem[Li et~al.(2022)Li, Li, Jiang, Weng, Geng, Li, Ke, Li, Cheng, Nie, and
  {others}]{li_yolov6_2022}
Chuyi Li, Lulu Li, Hongliang Jiang, Kaiheng Weng, Yifei Geng, Liang Li, Zaidan
  Ke, Qingyuan Li, Meng Cheng, Weiqiang Nie, and {others}.
\newblock {YOLOv6}: {A} single-stage object detection framework for industrial
  applications.
\newblock \emph{arXiv:2209.02976}, 2022.

\bibitem[Li et~al.(2016)Li, Zhang, and Liu]{li_ternary_2016}
Fengfu Li, Bo~Zhang, and Bin Liu.
\newblock Ternary weight networks.
\newblock \emph{arXiv:1605.04711}, 2016.

\bibitem[Lin et~al.(2019)Lin, Gan, and Han]{lin_defensive_2019}
Ji~Lin, Chuang Gan, and Song Han.
\newblock Defensive {Quantization}: {When} {Efficiency} {Meets} {Robustness}.
\newblock In \emph{Proceedings of the {International} {Conference} on
  {Learning} {Representations}}, 2019.

\bibitem[Liu et~al.(2019)Liu, Simonyan, and Yang]{liu_darts_2019}
Hanxiao Liu, Karen Simonyan, and Yiming Yang.
\newblock {DARTS}: {Differentiable} {Architecture} {Search}.
\newblock In \emph{Proceedings of the {International} {Conference} on
  {Learning} {Representations}}, 2019.

\bibitem[Liu et~al.(2022)Liu, Cheng, Huang, Xing, and
  Shen]{liu_nonuniform--uniform_2022}
Zechun Liu, Kwang-Ting Cheng, Dong Huang, Eric~P Xing, and Zhiqiang Shen.
\newblock Nonuniform-to-uniform quantization: {Towards} accurate quantization
  via generalized straight-through estimation.
\newblock In \emph{Proceedings of the {IEEE}/{CVF} {Conference} on {Computer}
  {Vision} and {Pattern} {Recognition}}, 2022.

\bibitem[Louizos et~al.(2019)Louizos, Reisser, Blankevoort, Gavves, and
  Welling]{louizos_relaxed_2019}
Christos Louizos, Matthias Reisser, Tijmen Blankevoort, Efstratios Gavves, and
  Max Welling.
\newblock Relaxed {Quantization} for {Discretized} {Neural} {Networks}.
\newblock In \emph{Proceedings of the {International} {Conference} on
  {Learning} {Representations}}, 2019.

\bibitem[Martin et~al.(2001)Martin, Fowlkes, Tal, and
  Malik]{martin_database_2001}
D.~Martin, C.~Fowlkes, D.~Tal, and J.~Malik.
\newblock A database of human segmented natural images and its application to
  evaluating segmentation algorithms and measuring ecological statistics.
\newblock In \emph{Proceedings of the {IEEE} {International} {Conference} on
  {Computer} {Vision}}, 2001.

\bibitem[Nagel et~al.(2020)Nagel, Amjad, Van~Baalen, Louizos, and
  Blankevoort]{nagel_up_nodate}
Markus Nagel, Rana~Ali Amjad, Mart Van~Baalen, Christos Louizos, and Tijmen
  Blankevoort.
\newblock Up or {Down}? {Adaptive} {Rounding} for {Post}-{Training}
  {Quantization}.
\newblock In Hal~Daumé III and Aarti Singh, editors, \emph{Proceedings of the
  {International} {Conference} on {Machine} {Learning}}, 2020.

\bibitem[Polino et~al.(2018)Polino, Pascanu, and Alistarh]{polino_model_2018}
Antonio Polino, Razvan Pascanu, and Dan Alistarh.
\newblock Model compression via distillation and quantization.
\newblock In \emph{Proceedings of the {International} {Conference} on
  {Learning} {Representations}}, 2018.

\bibitem[Qin et~al.(2020)Qin, Gong, Liu, Shen, Wei, Yu, and
  Song]{qin_forward_2020}
Haotong Qin, Ruihao Gong, Xianglong Liu, Mingzhu Shen, Ziran Wei, Fengwei Yu,
  and Jingkuan Song.
\newblock Forward and backward information retention for accurate binary neural
  networks.
\newblock In \emph{Proceedings of the {IEEE}/{CVF} {Conference} on {Computer}
  {Vision} and {Pattern} {Recognition}}, 2020.

\bibitem[Russakovsky et~al.(2014)Russakovsky, Deng, Su, Krause, Satheesh, Ma,
  Huang, Karpathy, Khosla, Bernstein, Berg, and
  Fei-Fei]{russakovsky_imagenet_2014}
Olga Russakovsky, Jia Deng, Hao Su, Jonathan Krause, Sanjeev Satheesh, Sean Ma,
  Zhiheng Huang, Andrej Karpathy, Aditya Khosla, Michael~S. Bernstein,
  Alexander~C. Berg, and Li~Fei-Fei.
\newblock {ImageNet} {Large} {Scale} {Visual} {Recognition} {Challenge}.
\newblock \emph{International Journal of Computer Vision}, 2014.

\bibitem[Salimans and Kingma(2016)]{salimans_weight_2016}
Tim Salimans and Diederik~P. Kingma.
\newblock Weight {Normalization}: {A} {Simple} {Reparameterization} to
  {Accelerate} {Training} of {Deep} {Neural} {Networks}.
\newblock In \emph{Proceedings of the {Conference} on {Neural} {Information}
  {Processing} {Systems}}, 2016.

\bibitem[Simonyan and Zisserman(2014)]{simonyan_very_2014}
Karen Simonyan and Andrew Zisserman.
\newblock Very {Deep} {Convolutional} {Networks} for {Large}-{Scale} {Image}
  {Recognition}.
\newblock In \emph{Proceedings of the {International} {Conference} on
  {Learning} {Representations}}, 2014.

\bibitem[Solodskikh et~al.(2022)Solodskikh, Chikin, Aydarkhanov, Song,
  Zhelavskaya, and Wei]{solodskikh_towards_2022}
Kirill Solodskikh, Vladimir Chikin, Ruslan Aydarkhanov, Dehua Song, Irina
  Zhelavskaya, and Jiansheng Wei.
\newblock Towards {Accurate} {Network} {Quantization} with {Equivalent}
  {Smooth} {Regularizer}.
\newblock In \emph{Proceedings of the {European} {Conference} on {Computer}
  {Vision}}, 2022.

\bibitem[Wang et~al.(2022{\natexlab{a}})Wang, Bochkovskiy, and
  Liao]{wang_yolov7_2022}
Chien-Yao Wang, Alexey Bochkovskiy, and Hong-Yuan~Mark Liao.
\newblock {YOLOv7}: {Trainable} bag-of-freebies sets new state-of-the-art for
  real-time object detectors.
\newblock \emph{arXiv:\\2207.02696}, 2022{\natexlab{a}}.

\bibitem[Wang et~al.(2022{\natexlab{b}})Wang, Dong, and Shan]{wang_repsr_2022}
Xintao Wang, Chao Dong, and Ying Shan.
\newblock {RepSR}: {Training} {Efficient} {VGG}-style {Super}-{Resolution}
  {Networks} with {Structural} {Re}-{Parameterization} and {Batch}
  {Normalization}.
\newblock In \emph{Proceedings of the {International} {Conference} on
  {Multimedia}}, 2022{\natexlab{b}}.

\bibitem[Xia et~al.(2023)Xia, Zhang, Wang, Tian, Yang, Timofte, and
  Van~Gool]{xia_basic_2023}
Bin Xia, Yulun Zhang, Yitong Wang, Yapeng Tian, Wenming Yang, Radu Timofte, and
  Luc Van~Gool.
\newblock Basic {Binary} {Convolution} {Unit} for {Binarized} {Image}
  {Restoration} {Network}.
\newblock \emph{Proceedings of the International Conference on Learning
  Representations}, 2023.

\bibitem[Yamamoto(2021)]{yamamoto_learnable_2021}
Kohei Yamamoto.
\newblock Learnable companding quantization for accurate low-bit neural
  networks.
\newblock In \emph{Proceedings of the {IEEE}/{CVF} conference on computer
  vision and pattern recognition}, 2021.

\bibitem[Zagoruyko and Komodakis(2017)]{zagoruyko_diracnets_2017}
Sergey Zagoruyko and Nikos Komodakis.
\newblock {DiracNets}: {Training} {Very} {Deep} {Neural} {Networks} {Without}
  {Skip}-{Connections}.
\newblock \emph{arXiv:1706.0038}, 2017.

\bibitem[Zeyde et~al.(2010)Zeyde, Elad, and Protter]{zeyde_single_2010}
Roman Zeyde, Michael Elad, and Matan Protter.
\newblock On {Single} {Image} {Scale}-up {Using} {Sparse}-{Representations}.
\newblock In \emph{Proceedings of the {International} {Conference} on {Curves}
  and {Surfaces}}, 2010.

\bibitem[Zhang et~al.(2021)Zhang, Zeng, and Zhang]{zhang_edge-oriented_2021}
Xindong Zhang, Hui Zeng, and Lei Zhang.
\newblock Edge-oriented {Convolution} {Block} for {Real}-time {Super}
  {Resolution} on {Mobile} {Devices}.
\newblock In \emph{Proceedings of the {International} {Conference} on
  {Multimedia}}, 2021.

\bibitem[Zhou et~al.(2016)Zhou, Wu, Ni, Zhou, Wen, and
  Zou]{zhou_dorefa-net_2016}
Shuchang Zhou, Yuxin Wu, Zekun Ni, Xinyu Zhou, He~Wen, and Yuheng Zou.
\newblock Dorefa-net: {Training} low bitwidth convolutional neural networks
  with low bitwidth gradients.
\newblock \emph{arXiv:1606.06160}, 2016.

\bibitem[Zhuang et~al.(2021)Zhuang, Tan, Liu, Liu, Reid, and
  Shen]{zhuang_effective_2021}
Bohan Zhuang, Mingkui Tan, Jing Liu, Lingqiao Liu, Ian Reid, and Chunhua Shen.
\newblock Effective training of convolutional neural networks with low-bitwidth
  weights and activations.
\newblock \emph{IEEE Transactions on Pattern Analysis and Machine
  Intelligence}, 2021.

\bibitem[Zoph and Le(2017)]{zoph_neural_2017}
Barret Zoph and Quoc~V. Le.
\newblock Neural {Architecture} {Search} with {Reinforcement} {Learning}.
\newblock In \emph{Proceedings of the {International} {Conference} on
  {Learning} {Representations}}, 2017.

\end{thebibliography}

\end{document}

\renewcommand{\theequation}{S\arabic{equation}}
\renewcommand{\thefigure}{S\arabic{figure}}
\renewcommand{\thetable}{S\arabic{table}}
\renewcommand{\thealgorithm}{S\arabic{algorithm}}

\renewcommand{\thesection}{S\arabic{section}}
\renewcommand{\thesubsection}{S\arabic{section}.\arabic{subsection}}
\renewcommand{\thesubsubsection}{S\arabic{section}.\arabic{subsection}.\arabic{subsubsection}}

\maketitle

\newcommand{\ik}[1]{\textcolor{magenta}{\parbox{\linewidth}{#1}}}

In supplementary material, we provide two sections.
The first one is devoted to the description of how we select hyperparameters for various algorithms and the motivation behind this selection.
The second one provides a derivation of our approximate form of \ac{bn}.

\section{Experimental setup}
We have reproduced the results of the \ac{fp} models using official repositories\footnote{Links to official repositories that we used: \href{https://github.com/xindongzhang/ECBSR}{ECBSR}, 
\href{https://github.com/JUGGHM/OREPA\_CVPR2022}{OREPA},
\href{https://github.com/DingXiaoH/RepVGG}{RepVGG}. \\
}, except for AC-ResNet-18, which we implemented from scratch in TensorFlow. 
% \ik{You can also cite the exact repositories here}
To ensure reproducibility, we provide the exact hyperparameters used for all experiments in Table ~\ref{table:params}.
We maintain the same number of epochs, batch size, loss function, optimizer, weight decay, and schedule for quantized training as in the \ac{fp} training. We only modify a \ac{lr}. We reduced the \ac{lr} by a factor of ten for the \ac{qat} stage in comparison with \ac{fp} training \ac{lr} for the classification models. 
% \ik{Reduced from which value?}
The RepQ method and the baselines are trained using the same \ac{lr}, except for the Merged 8-bit quantization baseline. For it, we additionally reduced \ac{lr} since it benefited the quality. For instance, the 8-bit RepVGG-A0 Merged baseline achieves a 68.65 accuracy with a \ac{lr} of 0.01 and a 69.21 accuracy with a \ac{lr} of 0.001.

Additional quantization parameters, referred to as steps, are introduced in \cite{esser_learned_2020}. 
We use independent steps for each channel of a weight tensor \cite{li_mqbench_2021}. We adjust the ResNet-18 steps' \ac{lr} to ensure training stability. For example, with steps' \ac{lr} of 0.01, the 4-bit OREPA-Resnet-18 model converges to an accuracy of 65.49, and with steps' \ac{lr} of 0.001, it achieves an accuracy of 71.49.

All aforementioned \ac{lr} adjustments are displayed in Table~\ref{table:params}.

We deliberately chose not to modify hyperparameters other than the \ac{lr} to facilitate a fair comparison between the proposed methods and the baselines. Selecting optimal hyperparameters for each experiment can be challenging and resource-intensive. However, quantized models can sometimes benefit from certain hyperparameter fine-tuning. For instance, reducing the weight decay by a factor of two for the 4-bit OREPA-Resnet-18 model improves its quality from 71.49 to 72.15. 
% Training for the full number of epochs might also be unnecessary, particularly for 8-bit quantization.

For 4-bit quantization, it is a common practice to keep the weights or inputs of the first and last layers in 8-bit, as these layers are more sensitive to quantization. In the case of 4-bit ECBSR and OREPA-Resnet-18, we keep the first and last layers fully 8-bit. For AC-Resnet-18, only the first layer is 8-bit, while for RepVGG, only the input of the first layers is 8-bit. Although there is a slight inconsistency in the configurations of different models, it does not affect the comparison of the methods and the resulting conclusions.

For BNEst experiments, we replace \ac{bn} layers with \ac{bn} estimation layers on both \ac{fp} and \ac{qat} stages.
For ResNet-18, we leave two last layers with \ac{bn}  due to the edge effect described in Section~\ref{BNE} and replace \ac{bn} with \ac{bn}
estimation in all other layers.
% \ik{present/past tense is not consistent}

Regarding model selection, we report the highest quality achieved on the ImageNet validation set during training for the classification models. For ECBSR, we report the results of the last checkpoint since this model is tested on multiple test sets.

\begin{table}[H]
    \small
    \begin{center}
    \begin{tabular}{ c  c  c  c  c }
        \hline
        Model 
        & \makecell{RepVGG-A0 \& \\ RepVGG-B0}
        & AC-Resnet-18
        & OREPA-Resnet-18
        & ECBSR \\
        \hline
        Epochs & 120 & 100 & 120 & 1000 \\ 
        % \hline
        Batch size & 256 & 256 & 256 & 32 \\
        % \hline
        Loss & \makecell{Cross entropy \\ + label smoothing 0.1} & Cross entropy &  Cross entropy & MAE \\ 
        % \hline
        Optimizer & \makecell{SGD \\ momentum 0.9}  & \makecell{SGD \\ momentum 0.9} & \makecell{SGD \\ momentum 0.9} & Adam \\ 
        % \hline
        Schedule & \makecell{Cosine + \\ 5 epochs \\ 
 linear warmup } & Cosine & \makecell{Cosine + \\ 5 epochs \\ 
 linear warmup } & Constant \\ 
        % \hline
        Weight decay & 0.0001 & 0.0001 & 0.0001 & 0.0 \\
        % \hline
        \makecell{FP models' \ac{lr}} & 0.1 &  0.1 & 0.1 & 0.0005 \\ 
        % \hline
        \makecell{Quantized \\ models' \ac{lr}} & 0.01 &  0.01 & 0.01 & 0.0005 \\ 
        % \hline
        \makecell{Merged 8-bit \\ models' \ac{lr}} & 0.001 &  0.001 & 0.001 & 0.0005 \\ 
        % \hline
        \makecell{ 8-bit steps' \ac{lr}} & 0.01 &  0.0001 & 0.0001 & 0.0005 \\ 
        % \hline
        \makecell{4-bit steps' \ac{lr}} & 0.01 &  0.001 & 0.001 & 0.0005 \\
        \hline
    \end{tabular}
    \end{center}
    \caption{Hyperparameters' setup.}
    \label{table:params}
\end{table}

% \begin{table}
% \scriptsize
% \begin{center}

% % \multirow{2}{*}{\text { Model }} & 
% % \multirow{2}{*}{\text { Method }} & \multicolumn{4}{c}{\text { Precision }} \\
% %  & & 32 \text { (FP) } & 8 & 4 & 4 \text { (progressive) } \\
% % \hline 
% $\begin{array}{cccccccccc}
% \hline
%  % \fontsize{9.5pt}
% \multirow{2}{*}{\text { Model }} &
% \multirow{2}{*}{\text {\makecell{\text{Epoch} \\ \text{number}}}} & 
% \multirow{2}{*}{\text {\makecell{\text{Batch} \\ \text{size  }}}} &
% \multirow{2}{*}{\text { Loss }} & 
% \multirow{2}{*}{\text { Optimizer }} & 
% \multirow{2}{*}{\makecell{\text{Weight} \\ \text{decay}}} &
% \multirow{2}{*}{\text { Schedule }} &  
% \multicolumn{3}{c}{\text { Learning rate }} \\
%  & & & & & & & \text{FP} & \makecell{\text{QRep} \\ \text{Plain}} & \text{Merged} \\
%   \hline
%   \text{ECBSR} & 1000 & 32 & \text{L1} & \text{Adam} & - & \text{Constant} & 0.0005 & 0.0005 & 0.0005 \\
  
%   \text{AC-ResNet-18} & 100 & 256 & \text{CE} & \text{SGD} & 0.0001 & \text{Cosine} & 0.1 & 0.01 & 0.001 \\

%   \text{RepVGG} & 120 & 256 & \makecell{\text{CE}\\ 0.1 \text{ smoothing}} & \text{SGD} & 0.0001 & \makecell{\text{Cosine} \\ 5  \text{ epoch} \\ \text{warmup} } & 0.1 & 0.01 & 0.001 \\
%   \hline
%   % 8 & TBD & TBD & TBD & TBD   & TBD &
%   % 4 & TBD & TBD & TBD & TBD   &  TBD\\
% \end{array}$
% \caption{Hyper-parameter setup}
% \label{table:params}
% \end{center}
% \end{table}

% Key things
% 1. All params table + explanation params may vary from in the article and repositories
% 2. Notice weight decay and 
% 3. Last layers setup 

\section{Batch Normalization Estimation}
\label{BNE}

\newcommand{\bigbracket}[1]{\Bigl [ #1 \Bigr ]}

Here we derive the Batch Normalization Estimation formulas provided in Section 4.3.
 
Let's introduce notation. Consider an arbitrary convolution operation, denoted as $X * W$, where $X$ is the input tensor with shape $[B, H, D, IN]$, and $W$ is the weight tensor with shape $[K_h, K_w, IN, OUT]$. In this notation, $B$ represents the batch size, $H$ is the height of the feature map, $D$ is the width of the feature map, $K_h$ is the height of the weight tensor, $K_w$ is the width of the weight tensor, $IN$ is the number of input channels, and $OUT$ is the number of output channels.

We define a flattening operator, denoted as $F$, which reshapes the tensor $X$ from shape $[B, H, D, IN]$ to shape $[B \cdot H \cdot D, IN]$.

Using this notation, we can express the convolution operation as a sum of several matrix multiplications,
\begin{equation}
    (X * W)^F = \sum_{
    \substack{
    0 \leqslant i < K_h \\
    0 \leqslant j < K_w
    }
    } 
    X[,i:H - K_h + 1 + i, j:D - K_w + 1 + j,]^{F} W_{i,j}.
\end{equation}

The notation $[i:j;\ldots]$ represents a slicing operator commonly used in Python to extract specific elements from an array or tensor.
 $W_{i, j} = W[i,j]$ is a matrix of shape $[IN, OUT]$. This decomposition is convenient for further deductions.

We define $\mathbb{E}$ as an operator that computes the sample mean over the batch, height and width dimensions of the tensor. $\mathbb{E}$ is defined for the flattened input similarly, so that $\mathbb{E}\bigbracket{X^{F}} = \mathbb{E}\bigbracket{X}$. The resulting tensor will have a shape $[OUT,]$. Similarly, the operator $\mathbb{V}$ calculates the sample variance over the same batch, height, and width dimensions. These statistics, denoted as $\mathbb{E}$ and $\mathbb{V}$, are essential components used in regular Batch Normalization techniques.

With these notations and operators defined, we can now proceed to derive the estimations for Batch Normalization statistics for arbitrary convolutions, starting with the mean,
\begin{equation}
\begin{gathered}
\mathbb{E} \bigbracket{X * W} = 
\mathbb{E} \bigbracket{(X * W)^{F}} = \\
\mathbb{E} \bigbracket{
    \sum_{
    \substack{
    0 \leqslant i < K_h \\
    0 \leqslant j < K_w
    }
    } 
    X[,i:H - K_h + 1 + i, j:D - K_w + 1 + j,]^{F} W_{i,j}
} = \\
    = \sum_{
    \substack{
    0 \leqslant i < K_h \\
    0 \leqslant j < K_w
    }
    } 
    \mathbb{E} \bigbracket{X[,i:H - K_h + 1 + i, j:D - K_w + 1 + j,]} W_{i,j}.
\end{gathered}
\end{equation}

We can make the assumption that $\mathbb{E} \bigbracket{X[, i: H - K_h + 1 + i, j: D - K_w + 1 + j,]} \approx \mathbb{E}\bigbracket{X}$, which implies that we can neglect the pixels near the edges of the feature map. This approximation error remains small when the weight tensor's shape is much smaller than the height and width of the feature maps, $K_h \ll H$, and $K_w \ll D$. In many networks, this condition holds true. For example, in state-of-the-art models, the ImageNet input image is regularly reshaped into 224x224, while the weights are typically 3x3 for most layers. However, it is worth noting that the height and width of the feature maps may be significantly reduced in the last layers, which can lead to less accurate estimates. That is why for ResNet-18 BNEst experiments, we left two last layers to use regular \ac{bn}.

By employing this approximation, we can proceed to deduce the final mean estimate,

% Edge effect
\begin{equation}
\begin{gathered}
\mathbb{E} \bigbracket{X * W} \approx \tilde{\mathbb{E}} \bigbracket{X * W} = 
\sum_{
\substack{
0 \leqslant i < K_h \\
0 \leqslant j < K_w
}}
\mathbb{E} \bigbracket{X} W_{i, j} = 
\mathbb{E} \bigbracket{X} 
\sum_{
\substack{
0 \leqslant i < K_h \\
0 \leqslant j < K_w
}}
W_{i, j}.
\end{gathered}
\end{equation}

Similarly, we provide equations for estimating the variance,

\begin{equation}
\begin{gathered}
    \mathbb{V} \bigbracket{X * W}  = \mathbb{V} \bigbracket{(X * W)^{F}} = \\
    \mathbb{V}\bigbracket{
    \sum_{
    \substack{
    0 \leqslant i < K_h \\
    0 \leqslant j < K_w
    }}
    X[,i:H - K_h + 1 + i, j:D - K_w + 1 + j ,]^{F} W_{i,j}
    }.
\end{gathered}  
\end{equation}
We further assume that the variance of the sum is approximately equal to the sum of the variances,
\begin{equation}
\begin{gathered}
    \mathbb{V} \bigbracket{X * W}
    \approx  
    \sum_{
    \substack{
    0 \leqslant i < K_h \\
    0 \leqslant j < K_w
    }} \mathbb{V}\bigbracket{X[,i:H - K_h + 1 + i, j:D - K_w + 1 + j,]^{F} W_{i,j}}.    
\end{gathered}
\end{equation}
Using equation (4) from the article,
\begin{equation}
\begin{gathered}
    \mathbb{V} \bigbracket{X * W}
    \approx  
    \sum_{
    \substack{
    0 \leqslant i < K_h \\
    0 \leqslant j < K_w
    }} \mathbb{V}\bigbracket{X[,i:H - K_h + 1 + i, j:D - K_w + 1 + j ,]} W_{i,j}^2.    
\end{gathered}
\end{equation}
Similarly to $\mathbb{E}$, we neglect the edge effect, assuming,

\begin{equation}
    \mathbb{V} \bigbracket{X[,i:H - K_h + 1 + i,j:D - K_w + 1 + j,]} \approx \mathbb{V} \bigbracket{X}.
\end{equation}
We get the final variance estimate,
\begin{equation}
\begin{gathered}
    \mathbb{V} \bigbracket{X * W}
    \approx  \tilde{\mathbb{V}} \bigbracket{X * W} =
    \mathbb{V}\bigbracket{X}
    \sum_{
    \substack{
    0 \leqslant i < K_h \\
    0 \leqslant j < K_w
    }}  W_{i,j}^2.    
\end{gathered}
\end{equation}

% \textbf{Discussion[May be delete].}
% \ik{Let's discuss it tmw}
% We also explore the impact of Batch Normalization Estimation~(BNEst) on convolutions with different shapes on full-precision RepVGG. The RepVGG re-parametrization involves two types of convolutions: one with the shape of 3x3 and another with the shape of 1x1. Based on our derivations, we expect that the BNEst is more accurate for 1x1 convolutions than for 3x3 convolutions. To test this hypothesis, we place BNEst only after the 1x1 convolutions, while the 3x3 convolutions are followed by regular Batch Normalization. The results of this experiment are presented in Table ~\ref{table:edge}.  This configuration does not lead to any training speed improvement. However, surprisingly, it performs better than pure Batch Normalization and Batch Normalization estimation methods.
% \input{tables/edge}

\bibliography{supp_egbib}
% \printbibliography